\documentclass{article}

 \usepackage[preprint]{neurips_2025}

\usepackage[utf8]{inputenc} 
\usepackage[T1]{fontenc}    
\usepackage{hyperref}       
\usepackage{url}            
\usepackage{booktabs}       
\usepackage{amsfonts}       
\usepackage{nicefrac}       
\usepackage{microtype}      
\usepackage{xcolor}         
\usepackage{times}
\usepackage{latexsym}
\usepackage{algorithm}
\usepackage{algorithmic}
\usepackage{pifont}
\usepackage{multirow}
\usepackage{subcaption}
\usepackage{graphicx}
\usepackage{hyperref}

\usepackage{todonotes}

\title{Scaling Competence, Shrinking Reasoning: Cognitive Signatures in Language Model Learning}

\author{Mukul Singh \\
  Microsoft, USA \\
  \texttt{singhmukul@microsoft.com} \\
  \And
  Ananya Singha \\
  Microsoft, India \\
  \texttt{ananyasingha@microsoft.com} \\
  \And
  Arjun Radhakrishna \\
  Microsoft, USA \\
  \texttt{arradha@microsoft.com} \\
  \And
  Sumit Gulwani \\
  Microsoft, USA \\
  \texttt{sumitg@microsoft.com} \\
  }

\begin{document}

\maketitle

\begin{abstract}
We analyze reasoning in language models during task-specific fine-tuning and
draws parallel between reasoning tokens--\textit{intermediate steps generated while solving
problems} and the human working memory.
Drawing from cognitive science, we align training dynamics with the Four Stages
of Competence: models initially produce incorrect outputs without reasoning,
then begin reasoning (but still fail), eventually reason effectively, and
finally solve tasks without explicit reasoning.
We find that reasoning token length expands as performance improves, peaks at
the stage of conscious competence, then declines as the model internalizes the
task.
Notably, after training, models retain performance even when reasoning is
removed—suggesting it scaffolded learning but is no longer needed.
This progression offers actionable insights: reasoning token dynamics can serve
as a signal for diagnosing training stage, identifying convergence, and guiding
early stopping.
We propose metrics to track this trajectory and argue that reasoning behavior is valuable for understanding and optimizing reasoning model training.
\end{abstract}


\section{Introduction}
Recent advances in large language models have been driven by the development of reasoning models that generate intermediate steps, or reasoning tokens, to solve complex tasks \cite{openai2024openaio1card, deepseekai2025deepseekr1incentivizingreasoningcapability, Lozhkov2024StarCoder2}. 
These tokens enable models to decompose problems into smaller, interpretable inferential steps, significantly improving performance across domains such as code generation \cite{li2022alphacode}, mathematical reasoning \cite{ahn2024largelanguagemodelsmathematical}, regex synthesis \cite{xu2025largereasoningmodelssurvey}, natural language understanding \cite{wang2022mconala}, and logical question answering \cite{bigscience2022bloom}.
Reasoning also provides insights into the model's internal decision-making process such as planning \cite{reasoning-as-planning} and answer discovery \cite{xu2025largereasoningmodelssurvey}.
Previous work is divided on the role reasoning plays on the model performance, with studies like \cite{mousavi2025garbageinreasoningout, stop-anthropomorphizing-reasoning-token} suggesting that there is no direct correlation.
Despite the polarizing evidence, a systematic understanding of how reasoning tokens evolve during training and what insights into model learning they provide remains underexplored.


While some works caution against anthropomorphizing reasoning traces in language models—arguing that superficial analogies to human thought may obscure critical architectural differences and lead to misleading conclusions \cite{stop-anthropomorphizing-reasoning-token}—we argue that cognitive science concepts remain valuable analytic tools. Rather than assuming cognitive equivalence, such frameworks can serve as abstractions that help interpret model behaviors, diagnose learning dynamics, and guide the design of better training regimes.

For instance, empirical findings in cognitive psychology suggest that requiring humans to explicitly articulate their reasoning can impair task performance, revealing a tension between conscious deliberation and automatic execution \cite{cog-sci-dont-reason-too-much}. A parallel phenomenon has been observed in language models: on certain tasks, explicit reasoning can decrease performance \cite{ghosal2025doesthinkinghelpunderstanding}. These similarities have inspired mechanisms such as chain-of-thought prompting \cite{wei2023chainofthoughtpromptingelicitsreasoning}, incentivized reasoning traces \cite{deepseekai2025deepseekr1incentivizingreasoningcapability}, and inference-time scaling of reasoning behavior \cite{liu2025inferencetimescalinggeneralistreward}—suggesting that cognitively inspired perspectives can offer both descriptive and prescriptive insights for model development.

In this work, we take a closer look at the reasoning traces evolution through the training process of a reasoning model and draw parallels with learning theories for optimizing model learning.
We discover emergent patterns indicating that the reasoning tokens expand, peak and then contract as the model is trained on a task. The model accuracy and reliance on reasoning during this training process gives insights into learning stages of the model. Based on these, we propose metrics to track learning stages



To evaluate this, we conduct reinforcement learning experiments across multiple reasoning tasks--- code generation, math problems, regex synthesis, and logical question answering. Our results reveal a consistent pattern: reasoning token length increases alongside performance during early training, peaks at an intermediate stage corresponding to conscious competence, and then decreases as the model internalizes the task, reflecting a shift toward unconscious competence. Furthermore, we demonstrate that after training, models retain task performance even when reasoning tokens are removed, indicating that explicit reasoning initially scaffolds learning but becomes less necessary once competence is acquired.

These findings suggest that reasoning token dynamics can serve as actionable metrics to diagnose training stages, detect convergence, and inform early stopping criteria.

In summary, we make the following contributions:

\begin{enumerate}
    \item We formalize reasoning tokens as a working memory analog in language models, examining if this is a cognitive or biological effect.
    \item We empirically show that reasoning follows predictable cycle peaking, and contracting—as models learn diverse reasoning tasks.
    \item We propose metrics to for learning stages, enabling efficient training of reasoning models.
\end{enumerate}

\section{Related Works}

Recent progress in language models has demonstrated the power of explicit reasoning—via techniques like chain-of-thought prompting, scratchpads, and intermediate program synthesis—to solve complex tasks that require multi-step inference \cite{wei2023chainofthoughtpromptingelicitsreasoning, deepseekai2025deepseekr1incentivizingreasoningcapability}. These methods generate intermediate "reasoning tokens" that serve both as a decomposition of the problem and as a form of internal scaffolding. While reasoning has been shown to improve accuracy and interpretability across domains like math \cite{cobbe2021trainingverifierssolvemath}, code \cite{cassano2022multipl-e}, and logic \cite{talmor2019commonsenseqaquestionansweringchallenge}, most prior work has focused on prompting or output analysis. In contrast, little attention has been paid to how reasoning behavior evolves during training, or how it relates to internal representations and skill acquisition.

Cognitive science provides a rich theoretical lens through which to interpret this process. Working memory—the system responsible for holding and manipulating information temporarily—has been shown to play a key role in human reasoning, learning, and transfer \cite{oberauer2019working}. Notably, experimental work shows that prompting humans to consciously explain their reasoning can degrade performance in some tasks \cite{yax2024studying}, suggesting that reasoning is both a cognitive aid and a transient phase in learning. The "Four Stages of Competence" framework \cite{wikipedia_four_stages_of_competence} further formalizes how learners move from unskilled and unaware to fluent and automatic, with conscious reasoning peaking midway through this progression. These ideas have historically informed symbolic and neural cognitive architectures, but have not been directly applied to modern language model training.

Our work bridges this gap by drawing a parallel between reasoning tokens in models and working memory in humans—not as a biological function, but as a cognitive abstraction that may emerge in any intelligent system. Prior work has explored interpretability, training diagnostics \cite{reasoning-as-planning}, and early stopping strategies \cite{chen2020learningstoplearningpredict}, but existing metrics rarely account for the internal deliberative behavior of models. We position reasoning token dynamics as a measurable correlate of cognitive development within models, offering a new class of metrics for identifying competence stages and optimizing training. This complements recent interest in aligning learning signals with human-like cognitive trajectories \cite{mousavi2025garbageinreasoningout}, and contributes to a growing body of work connecting AI systems with cognitive theory.

\section{Reasoning and Memory}

\begin{figure}
    \centering
    \includegraphics[width=\linewidth]{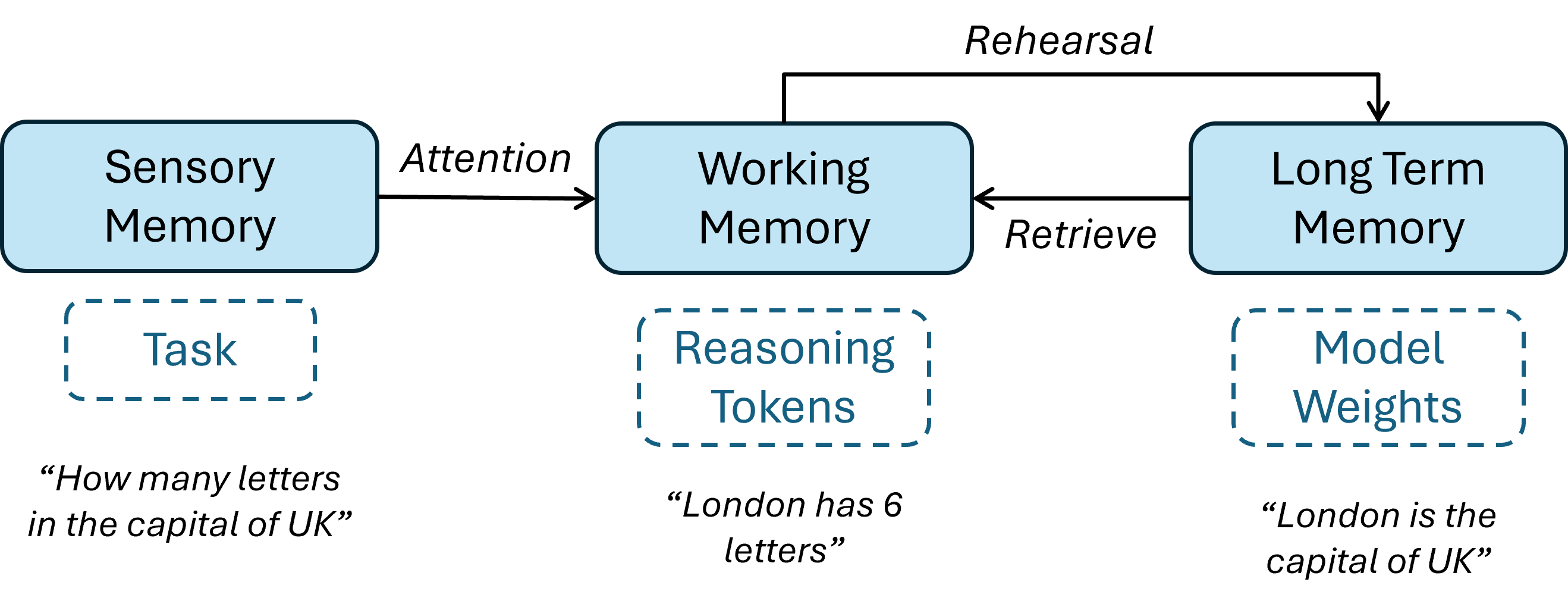}
    \caption{A map of cognitive memory structure.
    }
    \label{fig:cog-map}
\end{figure}

We describe the reasoning model setup and theories around memory and cognition.

\subsection{Four stages of competence}

There have been extensive studies on human cognition and knowledge acquisition with multiple frameworks. One such framework is called the four stages of competence which describe the cognition abilities while learning. This hypothesizes that learning any skill or task requires going through 4 stages: (1) unconscious incompetence: when a subject is bad at a task and is not aware, often overestimating its ability; (2) conscious incompetence: when a subject is bad at a task but is aware of its abilities and consciously tries to solve the task; (3) conscious competence: when a subject is able to solve tasks but it takes a lot of conscious effort; (4) when a subject has mastered a skill and does not need conscious effort to solve tasks.

We draw parallels between language model training and these 4 stages showing that the model training also predictably follows these paradigms.

\subsection{Reasoning as scaffold to distill knowledge}
Cognitive literature \cite{oberauer2019working} has established correlation between working memory as a scaffolding to distill information to the long term memory where high correlation between repetition in short term memory and distillation to long term memory is observed. The same concept can be extended to language models if reasoning tokens are theorized as the working (or short term) memory. What this means is that the model learns new information or skills through first internalizing the task in its working memory (conscious competence) and then eventually moving this knowledge to its long term memory (unconscious competence).

To capture this, we track the performance of the model both with and without reasoning tokens as a way to quantify its long-term memory through the hypotheses that the knowledge that exists in the long term memory, does not need reasoning. We also track the length of reasoning tokens to observe the extent in which working memory is utilized through the training process.

\subsection{Learning stages}
Reasoning tokens provide an insight into the working of language models.
We track stages of learning during the training process by analyzing the reasoning traces to identify patterns. In particular we measure: (1) the length of reasoning traces; (2) the accuracy of responses; (3) the correctness of reasoning traces; and (4) non-reasoning performance.
On observing the evolution of these metrics over the training process reveals common patterns that are model and domain agnostic.

\section{Experimental Setup}
We conduct experiments to support the learning hypotheses introduced in the paper. We use reinforcement learning as the primary training method since it is closest to task based knowledge acquisition. We use standard reinforcement learning strategies including (1) Proximal Policy Optimization (PPO) \cite{schulman2017proximalpolicyoptimizationalgorithms}; (2) Group Relative Policy Optimization (GRPO) \cite{shao2024deepseekmathpushinglimitsmathematical}; (3) Reinforce \cite{zhang2020sampleefficientreinforcementlearning}. We train for 100 iterations with 32 rollouts per iteration. We train on a cluster of 8xH100 and the overall training takes 2 hours 35 minutes on average per model.

\subsection{Benchmarks}
We consider three domains to validate the findings:
\begin{enumerate}
    \item \textbf{Code Generation}: we consider low resource programming languages Rust, Ocaml and PHP since the base performance for these languages is low. The task is MCQ over code generation, repair and understanding. The training reward is a boolean based on the correctness of the final answer.
    \item \textbf{Math Problems}: we use the GSM8K \cite{cobbe2021trainingverifierssolvemath} dataset of match problems. The training reward is a boolean based on the correctness of the final answer.
    \item \textbf{Reasoning}: we use the Commonsense QA benchmark \cite{talmor-etal-2019-commonsenseqa} with MCQ and word based logical questions. The final reward is a boolean based on the correctness of the final answer.
\end{enumerate}

\subsection{Metrics}
We track (1) the number of reasoning tokens generated; (2) the correctness of the final answer; (3) the correctness of the reasoning trace; (4) the correctness of the final answer without reasoning. We track these every 50 steps of training.

\subsection{Models and Configuration}
We report results on both small and large language models; Deepseek-distill-7B, Phi-4-13B, GPT-4o-mini, GPT-o4-mini. We use the default setup of the model with two variations in inference -- (1) with reasoning which is the standard mode; and (2) without reasoning where the reasoning tokens are turned off and the model behaves like a regular completion model.

\section{Results}
We answer the following research questions:
\begin{itemize}
    \item[\textbf{RQ1.}] Do language models show the four stages of competence?
    \item[\textbf{RQ2.}] Does reasoning act as a scaffolding for knowledge distillation?
    \item[\textbf{RQ3.}] Can the proposed metrics predict training stages?
\end{itemize}

\subsection{RQ1: Stages in Learning}
\begin{figure}
    \centering
    \includegraphics[width=0.7\linewidth]{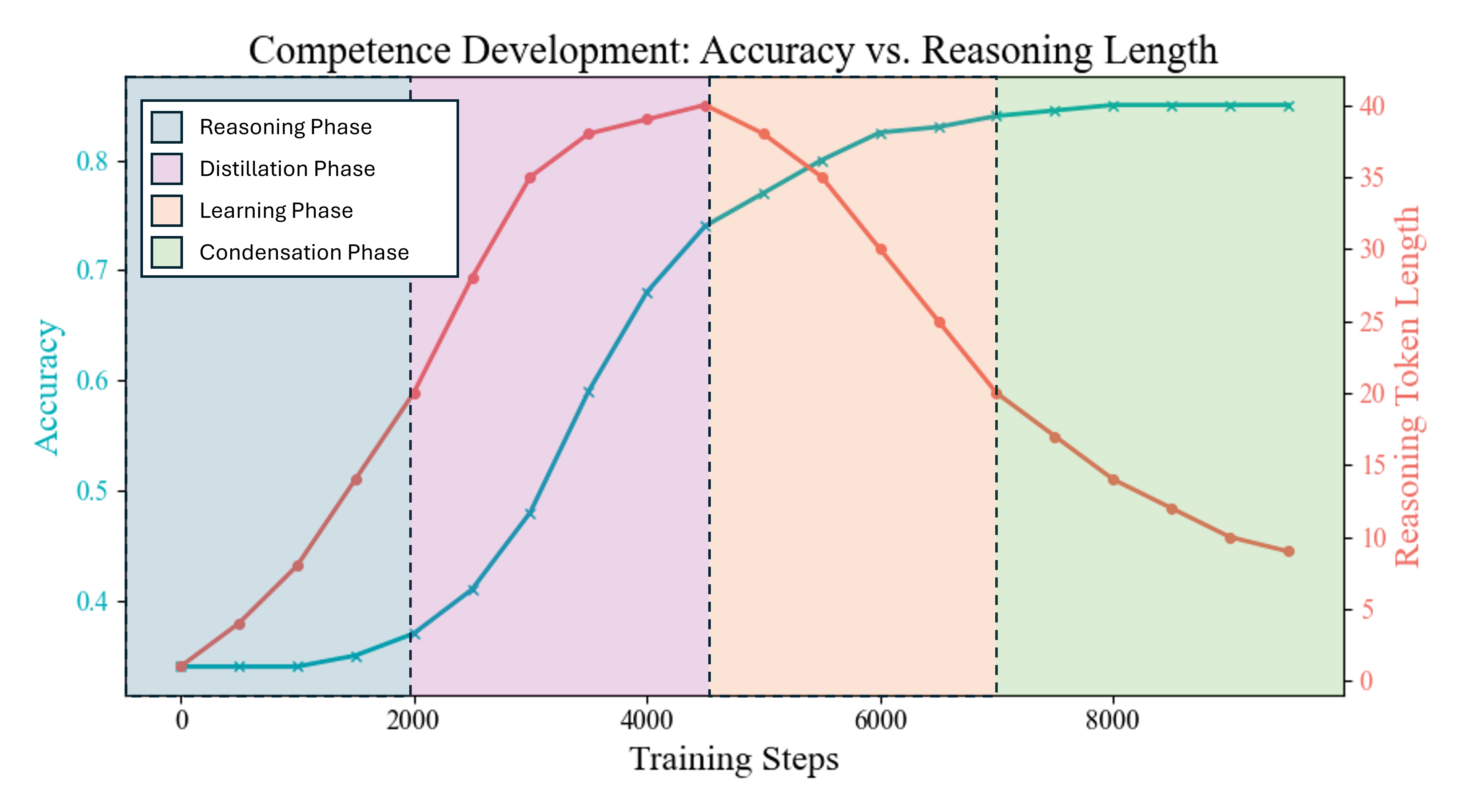}
    \caption{Distribution of the reasoning length (\# of tokens) with the task accuracy.}
    \label{fig:reasoning-vs-accuracy}
\end{figure}

We investigate the patterns in the reasoning and accuracy as the model is trained iteratively. We measure the length of reasoning tokens along with the final answer correctness. Figure~\ref{fig:reasoning-vs-accuracy} shows the distribution of the reasoning length (\# of tokens) with the task accuracy cumulated across multiple models and the tasks. 

We can see four clear regions in the plot which we call (1) Reasoning phase: where the models starts trying to reason to complete the task but the task accuracy has not yet improved; (2) Acquisition phase: where the task performance starts to improve using the reasoning effort; (3) Learning phase: where the model task performance peaks and the reasoning length stops increasing; (4) where the model task performance plateaus but the reasoning length declines.

We also observe that the same curve is observed across multiple models (1) \texttt{Phi-4-reasoning}, (2) \texttt{Deepseek-distill-23B}, (3) \texttt{GPT-o4-mini}, (4) \texttt{GPT-4o} and tasks across different domains (1) math; (2) logic and (3) code generation.


\subsection{RQ2: Reasoning as Scaffolding}
We look at the evolution of the reasoning traces as the model acquires knowledge. For this experiment, we analyze the model every 50 learning iterations and run inference on the checkpoint over the benchmark set. In particular we classify the reasoning and final answer as correct or incorrect. For the final answer correctness we rely on the ground truth for the math QA, test cases for regex generation and code generation. For the reasoning correctness, we check if the reasoning is (1) correct -- does not have any factually incorrect statement; (2) complete -- should have all the steps and not skip any steps.

\begin{figure}
    \centering
    \includegraphics[width=0.7\linewidth]{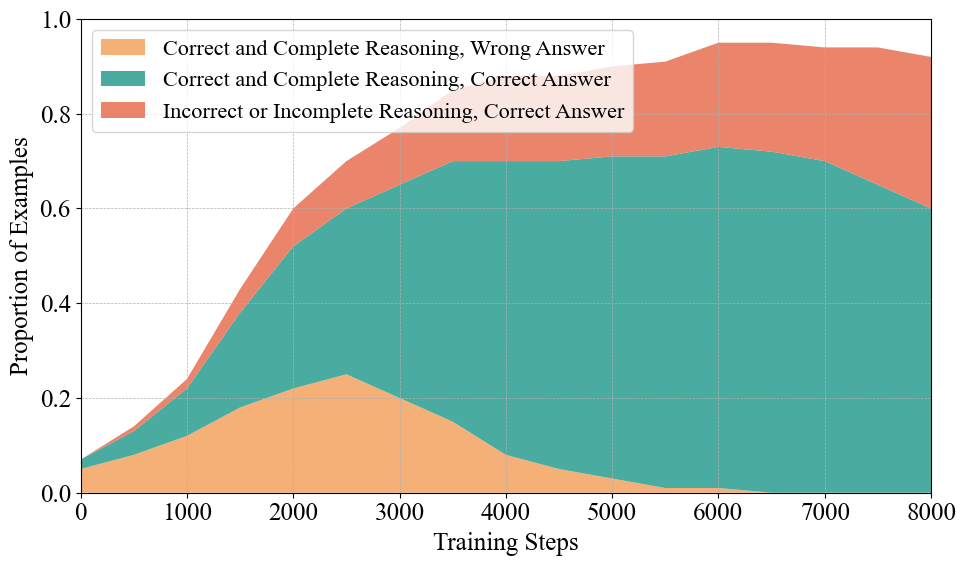}
    \caption{Distribution of reasoning and answer correctness across different training steps}
    \label{fig:reasoning-scaffolding}
\end{figure}

Figure~\ref{fig:reasoning-scaffolding} shows the distribution of reasoning and answer correctness with the training iterations. We see that the model accuracy increases through the correct reasoning accuracy goes down. 

\subsection{RQ3: Tracking Training Stages}
\begin{figure}
    \centering
    \includegraphics[width=0.9\linewidth]{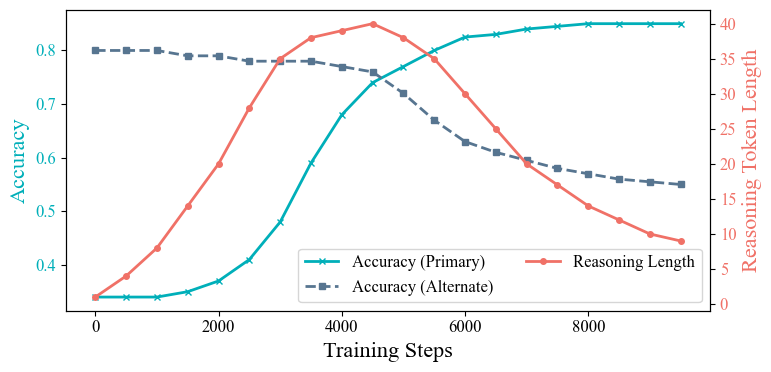}
    \caption{Performance of model on the training and held out datasets along with the length of reasoning length.}
    \label{fig:catastrophic-forgetting}
\end{figure}

Previous work has shown that reinforcement training is susceptible to catastrophic forgetting. We want to correlate the loss of knowledge with the learning stages. For this, we measure the performance of the model trained on one benchmark on the other two benchmarks. Figure~\ref{fig:catastrophic-forgetting} shows the model performance on the training and held out datasets along with the length of reasoning length.

We see that the model starts to rapidly lose information as soon as the reasoning peaks and starts to condense. Further, we see that the longer the model is trained, the reasoning condenses further and the performance on the held-out domains drops further.

\section{Conclusion}
Our analysis reveals that reasoning tokens serve as a dynamic scaffold during the fine-tuning of language models, echoing the progression described by the Four Stages of Competence in cognitive science. As models advance from incompetence to mastery, reasoning behavior emerges, peaks, and eventually recedes—highlighting a shift from explicit deliberation to internalized understanding. This trajectory not only deepens our understanding of model learning dynamics but also provides practical tools for diagnosing training progress and optimizing learning strategies. By treating reasoning token patterns as signals, we can better align training interventions with a model’s stage of competence, paving the way for more interpretable and efficient model development.

\section{Limitations}

Our work draws parallels between model reasoning behavior and cognitive science frameworks alongwith insights into training paradigms. There are several limitations and threats to validity.

\paragraph{Reasoning as proxy for cognition}
We treat reasoning token length and reasoning correctness as proxies for internal cognitive processes, but these metrics are still surface-level artifacts of text generation. The absence or brevity of reasoning tokens in late-stage models does not guarantee internalization in a human-like sense, and our use of competence stages is interpretive rather than diagnostic.

\paragraph{Domain specificity}
Our experiments are limited to a fixed set of tasks—primarily math, logic, and code generation. It remains an open question whether similar dynamics and competence transitions emerge in tasks with high ambiguity, contextual sensitivity, or real-world grounding (e.g., social reasoning or embodied tasks).

\paragraph{Learning environment}
While we propose cognitive-inspired metrics for training stage estimation, their utility may be less stable in settings with noisy supervision, curriculum learning, or large-scale instruction tuning where reasoning behavior is not cleanly separable.

Despite these limitations, we believe the cognitive lens remains a productive abstraction for studying and improving model training. Future work may formalize these analogies further or challenge them with deeper mechanistic analysis.




\bibliographystyle{abbrv}

\bibliography{custom}











\end{document}